\documentclass[10pt,twocolumn,letterpaper]{article}

\usepackage{iccv}
\usepackage{times}
\usepackage{epsfig}
\usepackage{graphicx}
\usepackage{amsmath}
\usepackage{amssymb}
\usepackage{subcaption}
\usepackage{multirow}
\usepackage{pifont}
\usepackage{floatrow}
\usepackage{tabularx}
\usepackage{bbm}
\usepackage{float}
\floatstyle{plaintop}
\restylefloat{table}
\usepackage[tableposition=top]{caption}

\usepackage[pagebackref=true,breaklinks=true,letterpaper=true,colorlinks,bookmarks=false]{hyperref}

\iccvfinalcopy 

\def\iccvPaperID{2729} 

\ificcvfinal\fi

\begin{document}

\title{Video Class Agnostic Segmentation with Contrastive Learning\\for Autonomous Driving}

\author{Mennatullah Siam\\
University of Alberta\\
Edmonton, Canada\\
{\tt\small mennatul@ualberta.ca}
\and
Alex Kendall\\
Wayve\\
London, UK\\
{\tt\small alex@wayve.ai}
\and 

Martin Jagersand\\
University of Alberta\\
Edmonton, Canada\\
{\tt\small jag@cs.ualberta.ca}
}

\maketitle
\ificcvfinal\thispagestyle{empty}\fi

\begin{abstract}
 Semantic segmentation in autonomous driving predominantly focuses on learning from large-scale data with a closed set of known classes without considering unknown objects. Motivated by safety reasons, we address the video class agnostic segmentation task, which considers unknown objects outside the closed set of known classes in our training data. We propose a novel auxiliary contrastive loss to learn the segmentation of known classes and unknown objects. Unlike previous work in contrastive learning that samples the anchor, positive and negative examples on an image level, our contrastive learning method leverages pixel-wise semantic and temporal guidance. We conduct experiments on Cityscapes-VPS by withholding four classes from training and show an improvement gain for both known and unknown objects segmentation with the auxiliary contrastive loss. We further release a large-scale synthetic dataset for different autonomous driving scenarios that includes distinct and rare unknown objects. We conduct experiments on the full synthetic dataset and a reduced small-scale version, and show how contrastive learning is more effective in small scale datasets. Our proposed models, dataset, and code will be released at~\url{https://github.com/MSiam/video_class_agnostic_segmentation}.
\end{abstract}

\section{Introduction}
\begin{figure}
    \centering
    \includegraphics[width=\textwidth]{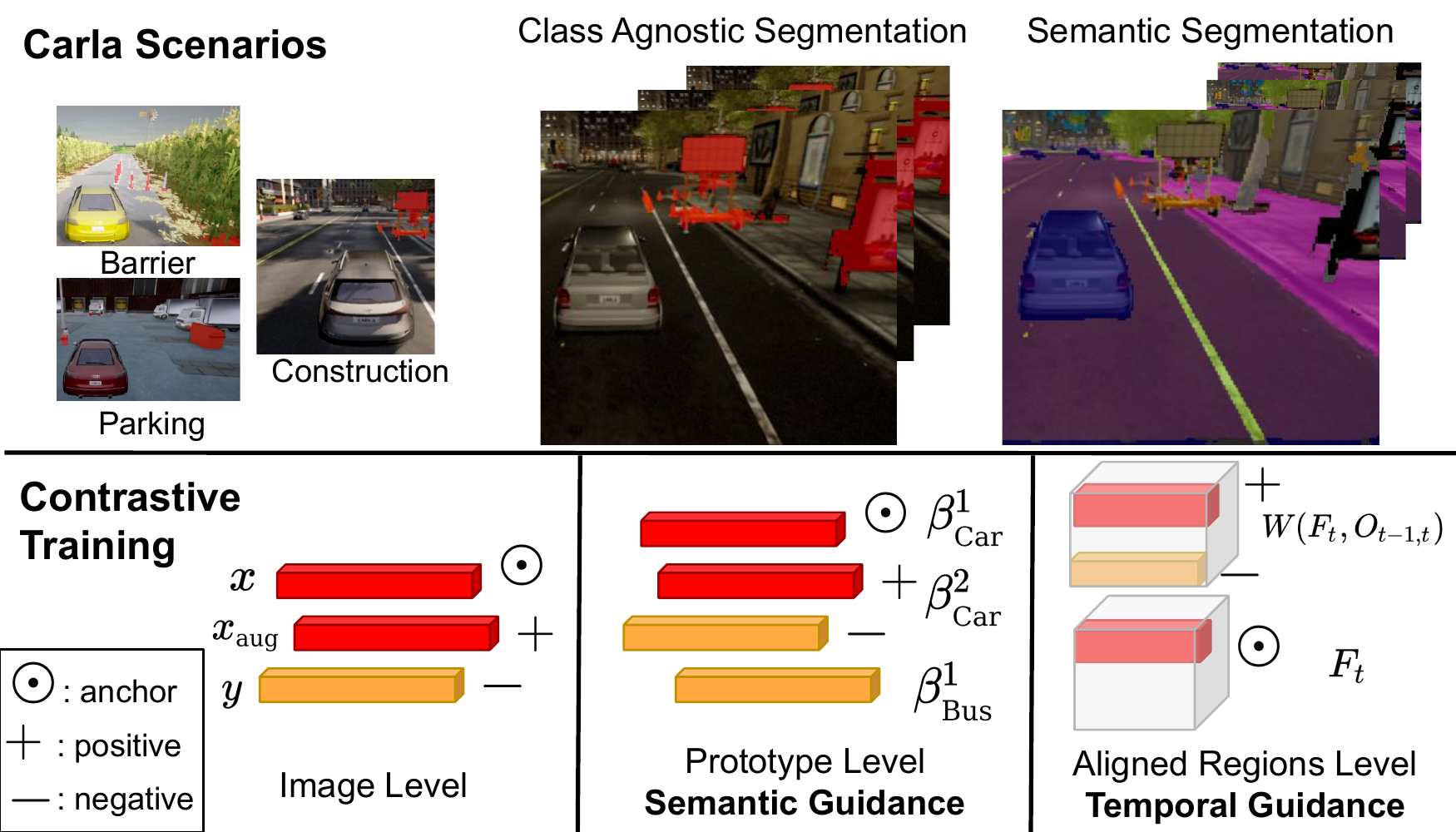}
    \caption{Video Class Agnostic Segmentation scenarios in Carla, showing example unknown classes, and our proposed auxiliary contrastive loss with semantic and temporal guidance.}
    \label{fig:teaser}
\end{figure}

Semantic scene understanding is crucial in autonomous driving in both end-to-end and mediated perception approaches as described in~\cite{chen2015deepdriving}. Semantic segmentation which performs pixel-wise classification of the scene is mostly trained on large scale data with closed set of known classes~\cite{cordts2016cityscapes}. However, a system trained on a limited set of classes would face difficulties in unexpected situations that could occur in different autonomous driving scenarios as shown in Figure~\ref{fig:teaser}. For example, construction and parking scenarios are not present in most semantic segmentation datasets~\cite{cordts2016cityscapes}, meaning that closed-set segmentation models will fail to detect multiple objects such as traffic warnings, garbage bins, and shopping carts.

To achieve this, we propose a novel contrastive learning method for video class agnostic segmentation which provides pixel-wise semantic and temporal guidance. Inspired by few-shot object segmentation literature~\cite{siam2019amp}\cite{zhang2019canet}\cite{wang2019panet}, we propose contrastive learning on the prototype-level with semantic guidance. Prototypes~\cite{wang2019panet} act as a class signature and are extracted using masked average pooling to represent a region for a certain class in the image. Since our task for unknown object segmentation in video sequences is concerned with pixel-wise embeddings, we perform contrastive learning on prototypes (representing different regions) unlike contrastive learning on the image-level~\cite{chen2020simple}. In our approach, we contrast a class prototype against the prototypes arising from other classes. Additionally, we propose a contrastive loss with temporal guidance to ensure the temporal consistency of embeddings through representation warping.

The closest work to our proposed contrastive learning variants are~\cite{winkens2020contrastive,khosla2020supervised,ke2021universal}. Contrastive training was proposed as a mean to improve out of distribution (OOD) detection in~\cite{winkens2020contrastive}, their proposed contrastive learning was on the image-level as they only handled classification. Their use for the contrastive loss was to ensure OOD detector robustness to variations in the imaging process such as lighting or camera position. A recent approach for contrastive learning explored supervised contrastive learning~\cite{khosla2020supervised}, but their work focused solely on image-level contrastive learning. Since our target task for segmenting unknown objects is a dense prediction task, we propose performing contrastive learning on the different regions using semantic and temporal guidance. The contrastive loss improves the discrimination among known and unknown objects/regions. In our experiments we ablate the different contrastive learning variants for image-level, prototype-level (semantic guidance) and temporally aligned regions-level (temporal guidance). 

To summarize, our main contributions are:

\begin{itemize}
    \item A novel method to learn video class agnostic segmentation using contrastive learning with semantic and temporal guidance. Unlike image-level contrastive learning~\cite{chen2020simple} our proposed variants are a better fit to pixel-level prediction tasks in video sequences.
    \item We provide different simulated driving scenarios in Carla \cite{dosovitskiy2017carla} and generate a large-scale synthetic dataset in autonomous driving for conducting controlled experiments on the video class agnostic segmentation task.
    \item Unlike previous work in identifying unknown objects in autonomous driving~\cite{wong2020identifying,ovsep20204d}, we provide an analysis of the relation among objects labelled as unknown during training and testing to asses the task difficulty.
\end{itemize}


\section{Related Work}

\paragraph{Unknown Objects Segmentation.}
Identifying unknown objects in autonomous driving has not been thoroughly studied in the literature with only two works to date~\cite{wong2020identifying,ovsep20204d}. Wong et. al.~\cite{wong2020identifying} proposed to learn prototypes for stuff and things (per instance) classes, inspired by prototypical networks for few-shot learning~\cite{snell2017prototypical}. A multi-frame bird's eye view representation from LIDAR pointclouds is used as input to their model. Secondly, Osep et. al.~\cite{ovsep20204d} proposed  a method that utilizes a video sequence of stereo images with 4D generic proposals for autonomous driving, utilizing parallax to identify temporally consistent objects. However, their method relies on a category-agnostic object proposal network which can ignore unknown objects that are not labelled in training examples. It has only been evaluated on 150 images for the open-set task in autonomous driving setting. It is also a computationally intensive method as they use a two-stage object detection, although most of the unknown objects are considered as stuff classes and does not require the separation of instances/proposals. We propose contrastive learning with semantic guidance to improve the unknown objects segmentation, our contribution is orthogonal to others and can work with multiple baseline architectures. Other related open-set detection and classification methods such as~\cite{dehghan2019online,winkens2020contrastive} are either constrained to certain downstream robotic tasks such as robot manipulation or focused on the image classification task. Both are much simpler than the segmentation task in diverse scenes for autonomous driving. 

\begin{figure*}[t!]
\begin{minipage}{0.3\linewidth}
\centering
\begin{tabular}{|l|l|}
\hline
             \textbf{Scenario}     & \textbf{Unknown Objects} \\ \hline
\multirow{3}{*}{Parking} & Cart with Bags \\
                  &  Shopping Trolley\\
                  &  Garbage Bin\\ \hline
\multirow{3}{*}{Construction} &  Traffic Warning\\
                  &  Construction Cone\\ \hline 
Barrier & Traffic Pole\\ \hline
\multirow{5}{*}{Training} &  Barrel\\
                          & Traffic Cone \\ 
                          & Traffic Barrier\\ 
                          & Static (others) \\
                          & Dynamic (others) \\ \hline
\end{tabular}%
\end{minipage}%
\begin{minipage}{0.68\linewidth}
\centering
\begin{subfigure}{.28\textwidth}
    \includegraphics[width=\textwidth]{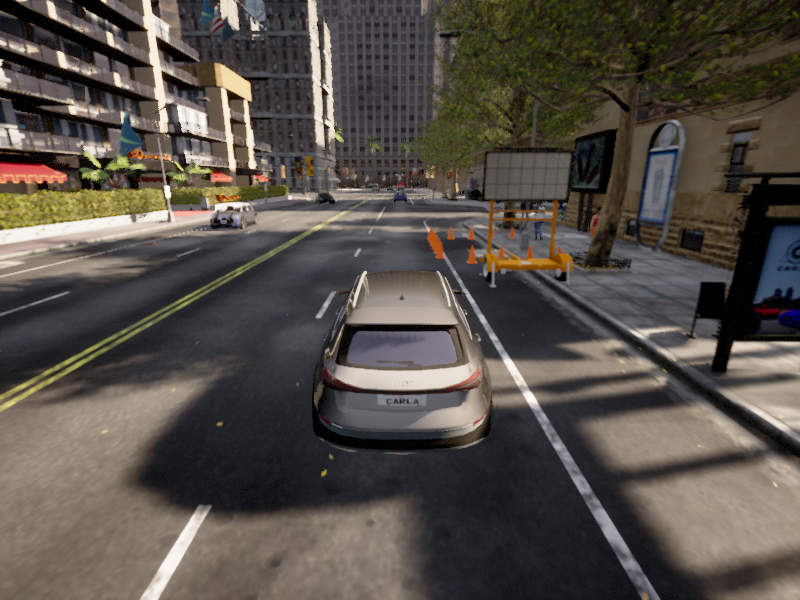}
\end{subfigure}%
\begin{subfigure}{.28\textwidth}
    \includegraphics[width=\textwidth]{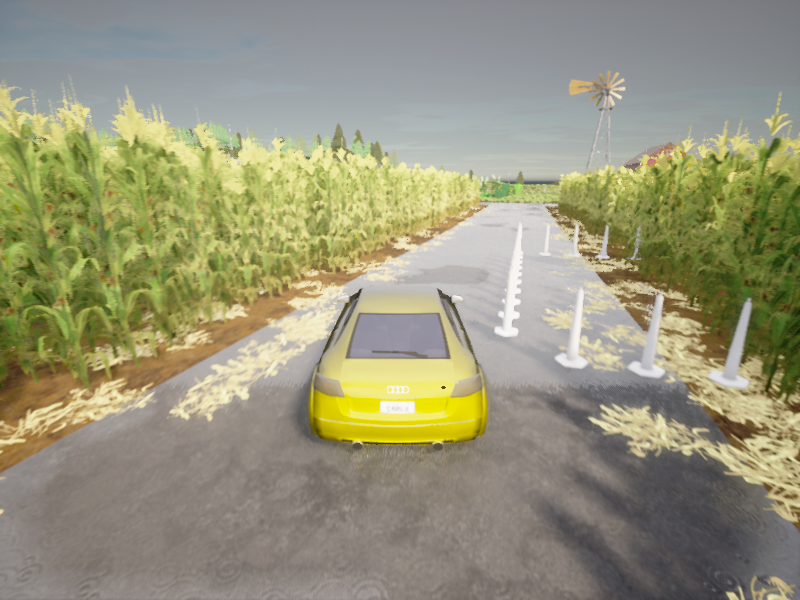}
\end{subfigure}%
\begin{subfigure}{.28\textwidth}
    \includegraphics[width=\textwidth]{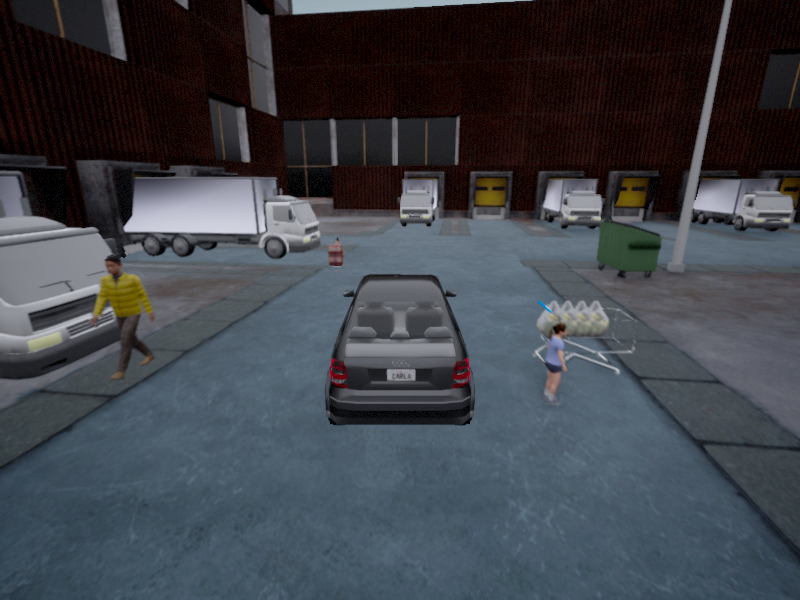}
\end{subfigure}

\begin{subfigure}{.28\textwidth}
    \includegraphics[width=\textwidth]{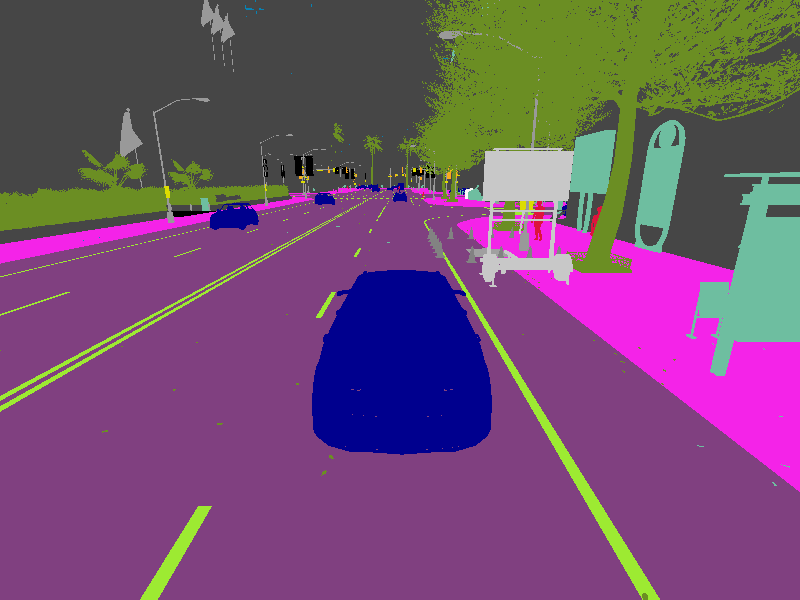}
\end{subfigure}%
\begin{subfigure}{.28\textwidth}
    \includegraphics[width=\textwidth]{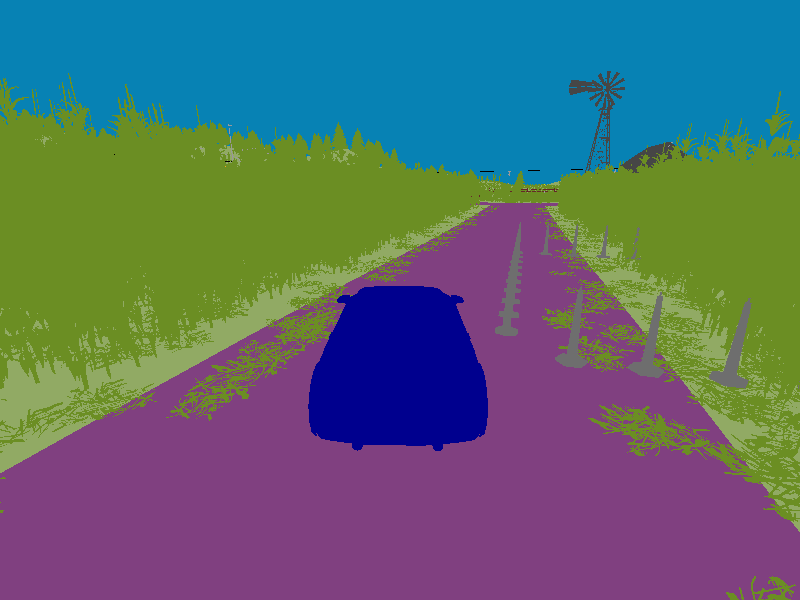}
\end{subfigure}%
\begin{subfigure}{.28\textwidth}
    \includegraphics[width=\textwidth]{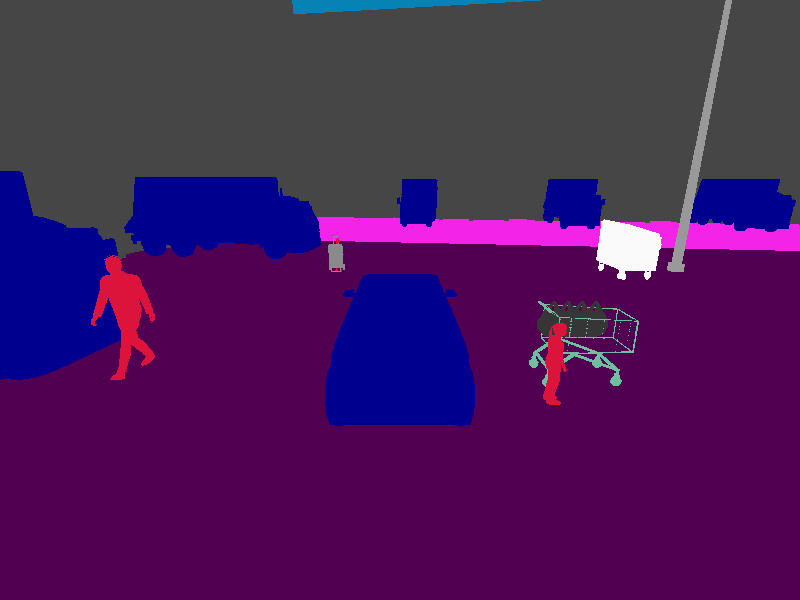}
\end{subfigure}
\end{minipage}
\caption{Our different scenarios in the Carla Simulation environment \cite{dosovitskiy2017carla} and objects considered as unknown in our synthetic data.}
\label{table:carla}
\end{figure*}

\paragraph{Contrastive Learning.}
Contrastive learning has recently been developed to boost unsupervised representation learning and improve transfer learning to other downstream tasks. Dosovitskiy et al.~\cite{dosovitskiy2014discriminative} developed the pretext task of instance discrimination to learn a representation in a self supervised manner using contrastive learning. There are two aspects to consider in contrastive learning: (1) supervised~\cite{khosla2020supervised,ke2021universal} versus unsupervised~\cite{chen2020simple,dosovitskiy2014discriminative}, i.e. whether labels are used to guide the contrastive learning process or not. (2) The contrastive learning mechanism, which may be end-to-end~\cite{chen2020simple}, using a memory bank~\cite{wu2018unsupervised}, or a momentum encoder~\cite{he2020momentum}. We compare supervised and unsupervised variants that are suitable to our downstream task with semantic (supervised) and temporal (unsupervised) guidance.

 Most of the literature focused on image-level contrastive learning~\cite{chen2020simple}~\cite{he2020momentum}. In a concurrent work Ke et. al.~\cite{ke2021universal} proposed contrastive learning on pixel-to-segment regions, since their downstream task was weakly supervised segmentation. In contrast, our task requires identifying unknown objects that can be leveraged with contrasting on the prototype-level, which improves the discrimination between known and unknown objects. For contrastive learning on videos, recent work~\cite{gordon2020watching} considered video sequences as a natural data augmentation. Because we are interested in temporally consistent embeddings for the segmentation task, we employ a different approach through representation warping with optical flow~\cite{gadde2017semantic}. We use temporally aligned region embeddings from consecutive frames as input to the contrastive loss.

\section{Video Class Agnostic Segmentation}
\section{Synthetic Dataset}
In this section we discuss the synthetic dataset that we curated to conduct controlled experiments on large-scale data to ensure the generalization ability of the class agnostic segmentation to unknown objects.

\subsection{Simulated Scenarios in Carla}
In the open-set segmentation formulation we care about providing video sequences along with annotations for unknown objects in different autonomous driving scenarios. Therefore, we build different scenarios within the Carla simulation environment~\cite{dosovitskiy2017carla}\footnote{We will incorporate these scenarios as part of the Carla challenge for autonomous driving~\url{https://leaderboard.carla.org/challenge/} to benefit both perception and policy learning researchers. Such scenarios serve as a way to evaluate the robustness of autonomous driving systems and how safety-critical are these systems designed.}. Additionally, the available video dataset with dense labels (Cityscapes-VPS~\cite{kim2020video}) has 3000 images for training, but only 300 for evaluation. Therefore, we collect our simulated data at large scale with 50K images for training and 2K for evaluation. In comparison, Wong et. al.~\cite{wong2020identifying} worked on LIDAR data and has not publicly released TOR-4D dataset, while Osep et. al.~\cite{ovsep20204d} evaluated on a small dataset of 150 images for unknown objects segmentation. 

Figure~\ref{table:carla} lists the three main scenarios that are used to evaluate the open-set segmentation task. We extend the Carla simulation environment to provide instance labels along with fine-grained labels for the specific set of unknown objects we introduce. The fine-grained labels are used to analyze the relation among unknown objects used during training and testing. We further modify the basic driving agent for the ego-vehicle to avoid unknown obstacles, in order to collect large-scale data. We randomize object placement, weather condition, traffic and use different unknown objects and towns between training and testing.

\begin{figure*}[t]
\centering
    \includegraphics[width=0.7\textwidth]{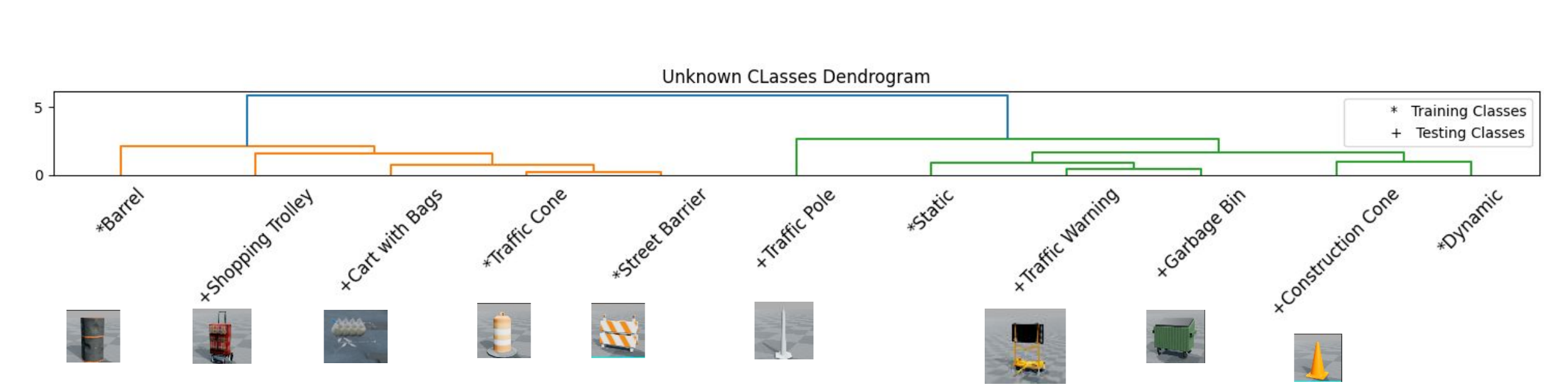}
    \caption{Dendrogram among unknown objects across training and testing data, showing their similarity computed in feature space.}
    \label{fig:dendrogram}
\end{figure*}

\subsection{Analysis of the Relationship between Unknown Objects}
The first question to assess in the open-set segmentation problem is to what extent the unknown objects that are used during the training phase are related to the ones used in testing. Answering this question can provide a better understanding of the task's difficulty and the open-set segmentation model's scalability. We propose to measure the distance between the prototype features for each instance, generated from masked average pooling with equation~\ref{eq:dist}. Where $F$ is the first two modules from ResNet-50 pretrained on Imagenet, $M$ is the semantic segmentation mask, $X_i$ $i^{th}$ input image with $N$ total number of images in the dataset and $P_c$ denotes the prototype for class $c$. Small scale data with fine-grained annotations for all unknown objects is collected and prototypes per class are computed then a pair-wise distance $d(i, j)$ is measured among classes. We use agglomerative clustering to form a dendrogram among the classes, as shown in Figure~\ref{fig:dendrogram}, to visualise and understand the relation between the unknown objects in the training and testing phases. 

Figure~\ref{fig:dendrogram} shows the classes that are visually similar, especially in texture, are semantically closer, such as street barrier and traffic cone. We also see less correlation among some of the different unknown objects used during testing and training. This strongly supports the claim that our model can generalise to distinct and unknown objects if it is successful in segmenting these unknown objects in our test data. Previous work~\cite{wong2020identifying,ovsep20204d} did not specify which pixels were labelled as unknown during training and testing, making it harder to evaluate the difficulty of their task. In our case we provide a way to visualise the relations among these objects in order to asses the task's difficulty.

\begin{subequations}
\begin{equation}
    P_c = \sum\limits_{i=1}^N \sum\limits_{x, y} \mathbbm{1}[M_i^{x, y}=c] F_{\theta}^{x, y} (X_i)
\end{equation}
\begin{equation}
    d(i, j) = \left\lVert P_i - P_j \right\rVert_2
\end{equation}
\label{eq:dist}
\end{subequations}

\section{Baseline Segmentation Network}
The baseline method we use for open-set segmentation has some similarities to~\cite{wong2020identifying}, where we learn a $\mu, \sigma$ to represent every class. Unlike~\cite{wong2020identifying} we focus on semantic segmentation, but our method can be easily extended to the panoptic segmentation task. Thus, our current baseline does not consider different instances for things classes. Our baseline model relies on appearance and geometry, as depth is a crucial signal for unknown object segmentation. Let the input appearance and depth be denoted as $x$, the segmentation label $y$ for $C$ classes, and $\hat{y}$ the predicted segmentation. 

We form a two-stream backbone model $f_{\theta}$ with ResNet-50 backbone~\cite{he2016deep} and a feature pyramid network~\cite{lin2017feature}. This is followed by a semantic segmentation head $f_{\phi}$ with 4 convolutional modules with ReLU and group normalization~\cite{wu2018group}, that learns $\mu, \sigma$ per class. The extracted features are $h = f_{\theta}(x)$, and the embeddings from the segmentation head are $m=f_{\phi}(h)$. The semantic segmentation head predicts the class probabilities following equation~\ref{eq:mahala}. Where the distance $d_{i, k}$ denotes the distance of pixel $i$ features to the representative signature of class $k$. A global trainable constant $\gamma$ is used to estimate the unknown objects that do not belong to any of the other known classes, regardless of the objects' semantics as in~\cite{wong2020identifying}. A softmax over $C+1$ distances is used to estimate the probability of the pixel to belong to a certain class.

\begin{subequations}
    \begin{equation}
        d_{i, k} = \frac{- \left\lVert m_i - \mu_k \right\rVert^2}{2 \sigma_k^2} 
    \end{equation}
    \begin{equation}
        d_{i, C+1} = \gamma
    \end{equation}
    \begin{equation}
        \hat{y}_{i, k} = \frac{\exp(d_{i, k})}{\sum\limits_{j=1}^{C+1}{\exp(d_{i, j})}}
    \end{equation}
    \begin{equation}
        l_{seg} = \frac{-1}{N} \sum\limits_{i=1}^N \sum\limits_{k=1}^{C+1}y_{i, k} \log{\hat{y}_{i, k}}
    \end{equation}
    \label{eq:mahala}
\end{subequations}

\section{Contrastive Learning}
In this section we detail our contrastive learning method. Our models have an input batch $\{ x, y\}_{i=1}^N$, and their pair $\{x', y'\}_{i=1}^N$. The other pair is determined based on semantic or temporal guidance being used as detailed in their respective sections. An extra contrastive learning head $f_{\alpha}$ is used to improve the separation of known classes and unknown objects. Similar to the segmentation head we use 4 convolutional modules with ReLU and group normalization~\cite{wu2018group}. The embeddings that are computed from the contrastive learning head are denoted as $z=f_{\alpha}(h)$.

\subsection{Semantic Guidance}
In order to improve the segmentation of known classes and more importantly the unknown objects we propose to perform contrastive learning on the prototype-level. We use semantic guidance to define a region as the segmentation mask of a certain class within the image, and extract prototypes using masked average pooling. The input to the model has a batch of images $X$ and the other pair $X'$ resulting in $2N$ images. We can use $X'$ as random augmentation on $X$, but since we use semantic segmentation to guide the extraction of different regions, the paired images do not necessarily have to be paired views. Rather it can be different images and it can serve as a natural augmentation, the segmentation mask will then guide the extraction of the corresponding regions. 
\begin{figure*}[t!]
\centering
    \includegraphics[width=0.8\textwidth]{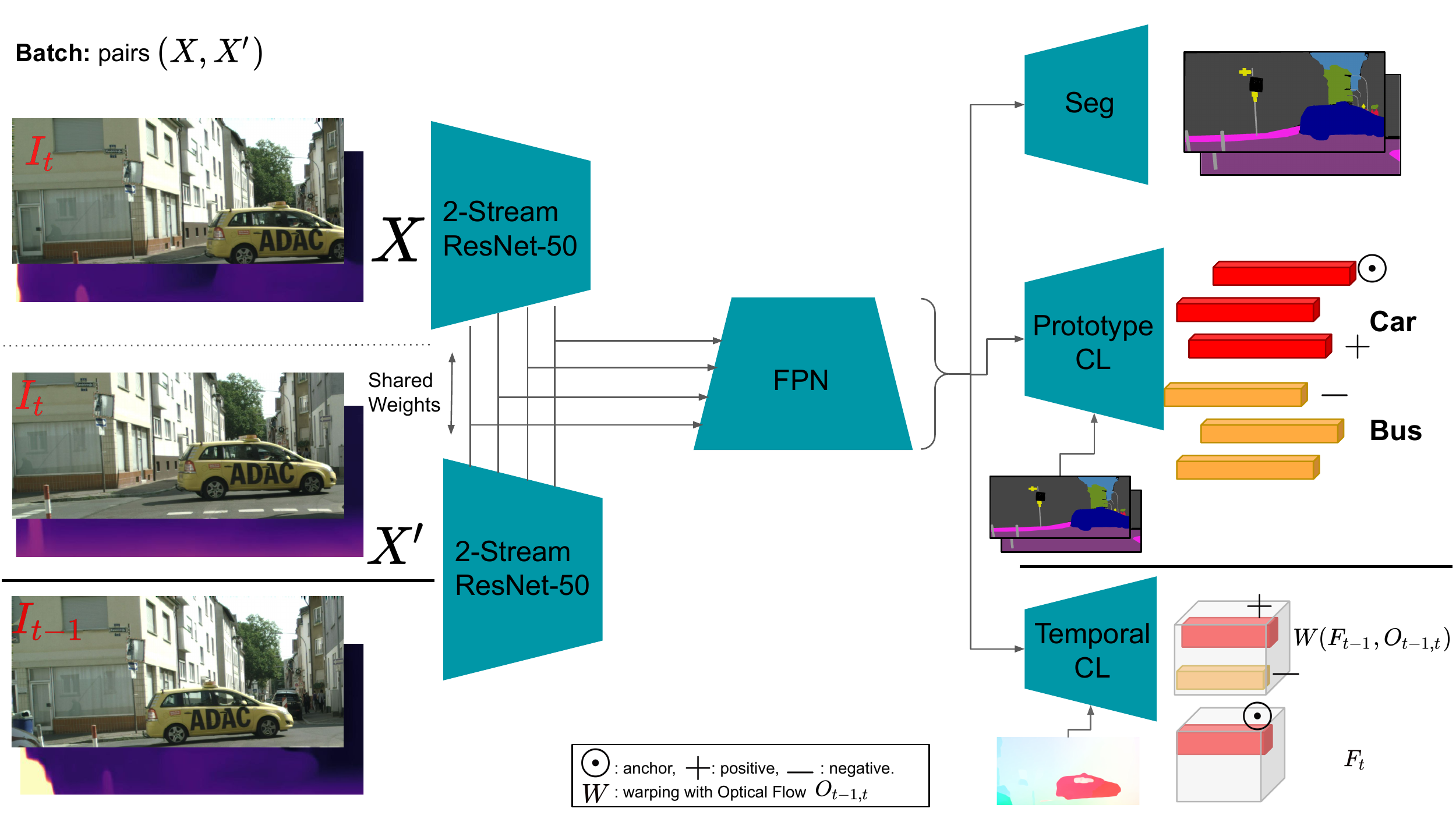}
    \caption{Our method for Video Class Agnostic Segmentation using Contrastive Learning. Input: appearance and depth. Prototype CL: semantic segmentation guides the region extraction, where positives are sampled from same class as the anchor, while negatives are from different classes. Temporal CL: aligns representation from $I_{t-1}$ to $I_t$ using Optical flow, then positive region is sampled as the aligned region to the anchor, while negatives are other regions in order to ensure temporal consistency of the embeddings.}
    \label{fig:model}
\end{figure*}

The contrastive learning with semantic guidance follows the equations in~\ref{eq:scl}, where $\beta_c$ is the output from masked average pooling on feature embeddings $z$ to represent class $c$ region. This results in a set of pairs $\{\beta_i, y_i\}_{i=1}^B$ with $y_i$ the corresponding class of the prototype $\beta_i$. This set is sampled from a memory queue with the previous batches features and from the current batch. In our model design of a two-stream encoder we can not use a momentum encoder~\cite{he2020momentum} due to practical limitations on the memory. Thus, we opt to using a memory queue which preserves the latest iterations' features with a limited memory size to reduce inconsistency among the features. In equation~\ref{eq:scl} $P(i)$ is the indices of the positive regions as $P(i) = \{p \in A(i): y_p = y_i\}$ and $A(i)$ is the set of indices of all prototypes. The final loss follows equation~\ref{eq:loss}.

\begin{subequations}
\begin{equation}
    \beta_c = \sum\limits_{i=1}^N \sum\limits_{x, y} \mathbbm{1}[M_i^{x, y}=c] z^{x, y}
\end{equation}
\begin{equation}
    L_{pcl} = \sum\limits_{i=1}^{B} \frac{-1}{|P(i)|} \sum_{p \in P(i)} \log{ \frac{\exp{(\beta_i. \beta_p / \tau)}}{\sum\limits_{a \in A(i)} \exp(\beta_i.\beta_a / \tau)} }
\end{equation}
\label{eq:scl}
\end{subequations}

\begin{equation}
    L = L_{seg} + \lambda L_{pcl}
    \label{eq:loss}
\end{equation}

\subsection{Temporal Guidance}
We propose another variant that employs temporal relations among embeddings from a video sequence in an unsupervised manner without the need of semantic segmentation labels. It is based on the concept of representation warping with optical flow to ensure temporal consistency of embeddings~\cite{gadde2017semantic}. Previous work in contrastive learning worked mainly on the video-level and was not concerned with the pixel-wise embeddings. However, since our main downstream task is segmentation we rather use representation warping with the estimated optical flow between two consecutive frames. Then compute the contrastive loss on the aligned regions from both frames belonging to a video sequence. Thus $X'$ pair is sampled as the temporally related frames to $X$ with the same random augmentation applied to both.

Temporal contrastive learning follows the equations in~\ref{eq:tcl}, where $F_t, F_{t-1}$ are the features extracted from consecutive frames and $O_{t, t-1}$ is the optical flow vector between them estimated using FlowNet2~\cite{ilg2017flownet}. $W$ is the warping operation resulting in $F'_{t-1}$, and $\text{AP}$ is the average pooling operation to combine the dense features into a smaller grid of regions $\delta$. In the contrastive loss the positive is sampled as the aligned region, and negatives are the remaining regions in the warped feature grid. The final loss follows equation~\ref{eq:loss2}

\begin{subequations}
\begin{equation}
F'_{t-1} = W(F_{t-1}, O_{t-1,t})
\end{equation}
\begin{equation}
\delta_t = \text{AP}(F_{t}), \delta'_{t-1} = \text{AP}(F'_{t-1})
\end{equation}
\begin{equation}
L_{tcl} = \sum\limits_{x,y} - \log{\frac{ \exp{( \delta_t(x, y). \delta'_{t-1}(x, y)}/ \tau)}{\sum\limits_{x', y'} \exp{(\delta_t(x, y). \delta'_{t-1}(x', y') / \tau})} }
\end{equation}
\label{eq:tcl}
\end{subequations}

\begin{equation}
    L = L_{seg} + \lambda L_{tcl}
    \label{eq:loss2}
\end{equation}

\section{Experimental Results}
\begin{table*}[t]
\centering
\caption{Open-Set Segmentation Results on Cityscapes-VPS and Carla. Fully supervised: Training the segmentation head on all cityscapes classes without a learnable global constant for the unknown object. CA-IoU: class agnostic IoU on the unknown objects.}
\label{table:vca}
\begin{tabular}{|l|l|l|l|c|c|}
\hline
Method & Dataset & Batch & Pretrained Weights & mIoU & CA-IoU  \\ \hline
Fully Supervised & Cityscapes~\cite{cordts2016cityscapes} & 4 & Supervised ImageNet & \textbf{65.5} & - \\ \hline
No CL & \multirow{2}{*}{Cityscapes-VPS~\cite{kim2020video}} & \multirow{2}{*}{4} & Supervised ImageNet & 63.2 & 17.9  \\
Prototype CL & & & Supervised ImageNet & \textbf{63.7} & \textbf{18.7} \\\hline
No CL & \multirow{5}{*}{Cityscapes-VPS~\cite{kim2020video}} & \multirow{5}{*}{2} & Supervised ImageNet & 56.4 &  18.4 \\
No CL & & & SimCLR\cite{chen2020simple} ImageNet & 56.1 & 17.3\\ 
Image CL & & & Supervised ImageNet & 60.1 & 19.8 \\ 
Prototype CL & & & Supervised ImageNet & \textbf{62.7} & \textbf{21.5}\\ 
Temporal CL & & & Supervised ImageNet & 61.7 & 21.4\\ \hline
No CL & \multirow{2}{*}{Carla} & \multirow{2}{*}{2} & Supervised ImageNet & \textbf{45.7} & \textbf{41.9} \\
Prototype CL & & & Supervised ImageNet & 44.2 & 37.2 \\\hline
\end{tabular}
\end{table*}

\subsection{Experimental Setup}
\textbf{Datasets:} We predominantly use Cityscapes-VPS~\cite{kim2020video} because it has dense annotations for the video sequences which is necessary when testing the temporal guidance aspect and our collected dataset on Carla. The dataset has 3000 images that are split into 2400 training images and 300 validation set images. Since the test set we do not have its annotations we rather ablate on the 300 validation set. Four classes from the 19 Classes are withheld to be considered as unknown, these are Person, Rider, Motorcycle, Bicycle. During training, we label Person, Rider, and some of the ignored classes in Cityscapes as unknown objects for training the global constant $\gamma$ and ignore pixels belonging to Motorcycle and Bicycle in the cross entropy loss. In the inference phase we rather use Motorcycle and Bicycle as the unknown objects and ignore the rest.

\textbf{Training Details:} Throughout all experiments we use SGD with momentum optimizer with 0.005 learning rate and 0.9 momentum, and weight decay of $1 \times 10^{-4}$. Following other segmentation methods~\cite{chen2017rethinking} we use a polynomial learning rate scheduling and train for 120,000 iterations on Cityscapes-VPS. We further found that a step learning rate scheduling improves the baseline without contrastive learning, where we reduce the learning rate with 0.1 at iterations 72,000, 96,000. We resize the images to $1024 \times 512$ then use random augmentations as random scales $\{0.8, 1.3\}$, random flipping and random cropping with $320 \times 512$ as crop sizes. We report mean intersection over union for the known classes. As for the unknown objects we report class agnostic IoU (CA-IoU), which computes the intersection over union on the unknown objects. CA-IoU is only reported on Motorcycle and Bicycle in Cityscapes-VPS since they are not previously seen during training, and in Carla the unknown objects for testing phase as shown in Figure~\ref{fig:dendrogram}. We freeze the initial layers for ResNet-50 up to the second module, and finetune the rest along with the feature pyramid network. In Carla we use the simulation groundtruth depth, and in Cityscapes-VPS we use the estimated depth from~\cite{godard2019digging}. We use a 1080Ti GPU for training most of the models.

\subsection{Results}
\subsubsection{Video Class Agnostic Segmentation Results}
In Table~\ref{table:vca} we evaluate our baseline segmentation network trained with all classes on Cityscapes, to evaluate the baseline performance on the general segmentation task without unknown objects. Then we start with conducting experiments on Cityscapes-VPS, and compare four main variants: (1) No contrastive learning (CL) baseline. (1) Image level (CL) after performing global average pooling on the features. (2) Prototype level with semantic guidance. (3) Aligned regions level with temporal guidance. Table~\ref{table:vca} shows the results for two sets of experiments with batch size 4 and 2 on Cityscapes-VPS. Due to limited GPU memory we were not able to conduct experiments for the temporal guidance variant using the larger batch size. However, the effective batch size used in the contrastive loss is different. A batch size of 2 will result in an effective batch size of 35 on average in the prototype-level variant, which are the prototypes extracted in an image along with the ones form the memory queue. As for the temporal variant it will result in 100, which are the dimensions (W.H) of the feature map output from average pooling. Batch size 4 experiments were conducted using polynomial learning rate scheduling as followed by~\cite{chen2017rethinking}, while batch size 2 set of experiments were conducted using step learning rate scheduling which we found to improve the baseline further. Further experiments on batch 4 with step learning rate scheduling and an improved baseline are shown in appendix~\ref{improved_base}.

In both sets of experiments the prototype-level CL improves the CA-IoU. The temporal CL improves over the baseline, but does not outperform the prototype-level variant. However, the temporal variant has the advantage that it does not need any labelled data unlike the prototype-level variant. Our experiments show as well that both semantic and temporal guidance improves over the use of image-level CL. Since our main task is video segmentation it is not sufficient to contrast embeddings globally. In summary, our initial experiments show that the segmentation task for both known and unknown classes benefits from the auxiliary contrastive loss especially with semantic guidance. Since semantic guidance improves the discrimination between known and unknown classes, while temporal guidance ensures the temporal consistency of the embeddings. 

\begin{figure*}[t]
\begin{minipage}{0.2\linewidth}
\centering
\begin{tabular}{|l|c|}
\hline
  Scenario   & CA-IoU  \\ \hline
 Barrier & 22.1\\
 Construction & 41.4 \\
 Parking & 27.0\\\hline
\end{tabular}
\end{minipage}%
\begin{minipage}{0.75\linewidth}
\centering
\begin{subfigure}{.3\textwidth}
    \includegraphics[width=\textwidth]{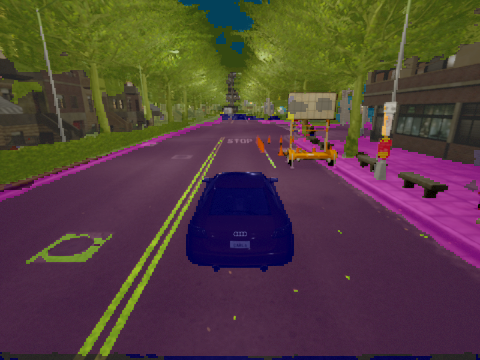}
\end{subfigure}%
\begin{subfigure}{.3\textwidth}
    \includegraphics[width=\textwidth]{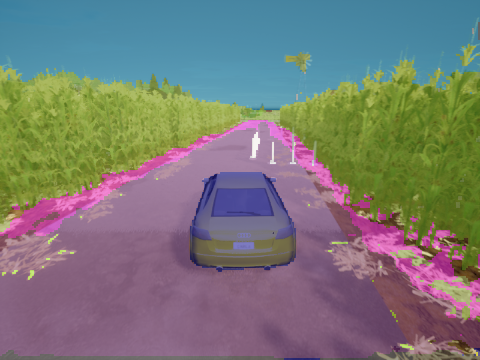}
\end{subfigure}%
\begin{subfigure}{.3\textwidth}
    \includegraphics[width=\textwidth]{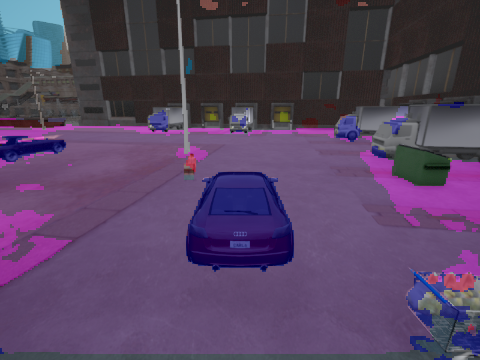}
\end{subfigure}

\begin{subfigure}{.3\textwidth}
    \includegraphics[width=\textwidth]{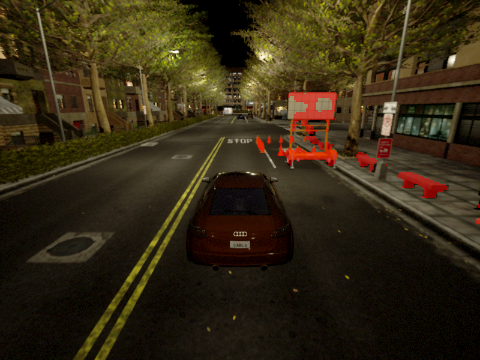}
    \caption{}
\end{subfigure}%
\begin{subfigure}{.3\textwidth}
    \includegraphics[width=\textwidth]{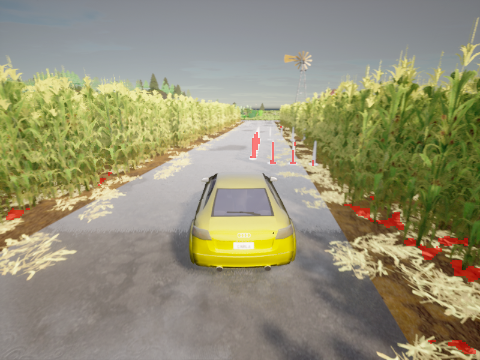}
    \caption{}
\end{subfigure}%
\begin{subfigure}{.3\textwidth}
    \includegraphics[width=\textwidth]{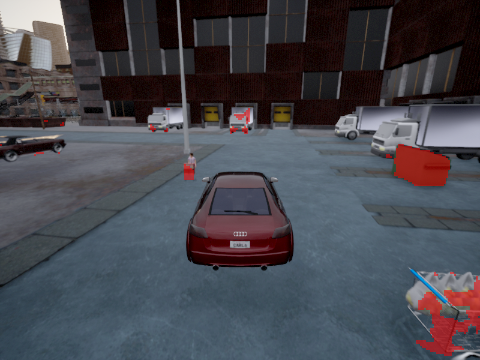}
    \caption{}
\end{subfigure}
\end{minipage}
\caption{CA-IoU reported per scenario. Predicted semantic and class agnostic segmentation on Carla Scenarios (a) Construction. (b) Barrier. (c) Parking. Top: semantic segmentation. Bottom: class agnostic segmentation}
\label{fig:opensetpred_carla}
\end{figure*}

\begin{figure*}[t]
\centering
\begin{subfigure}{.33\textwidth}
    \includegraphics[width=\textwidth]{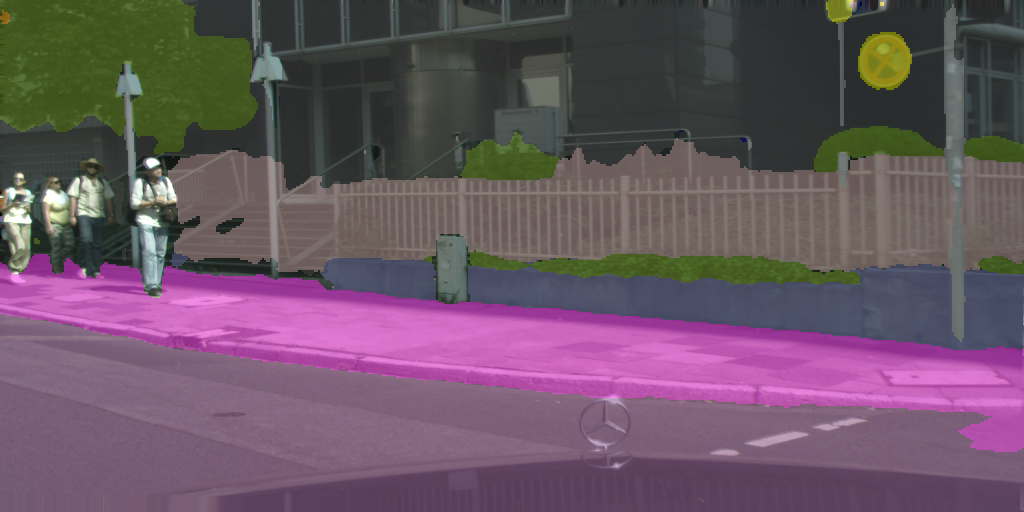}
\end{subfigure}%
\begin{subfigure}{.33\textwidth}
    \includegraphics[width=\textwidth]{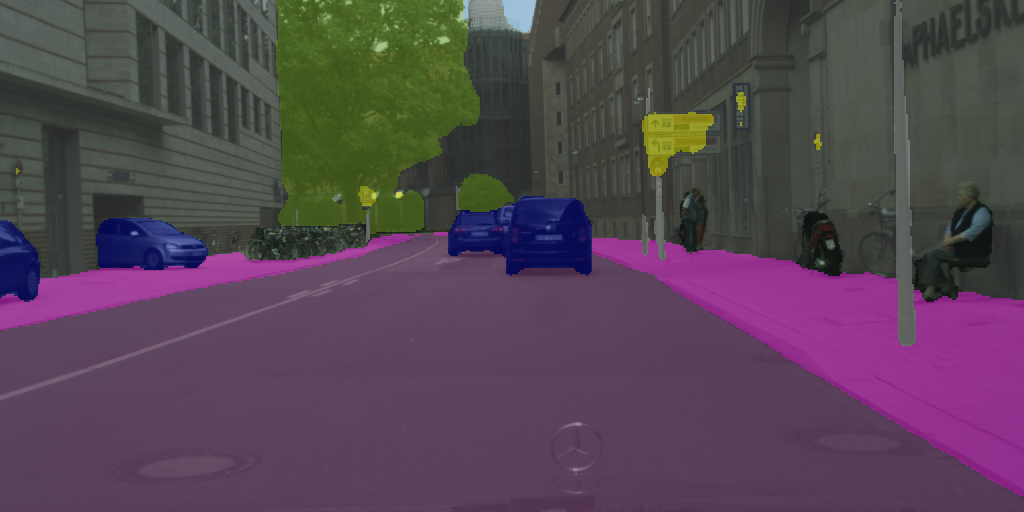}
\end{subfigure}%
\begin{subfigure}{.33\textwidth}
    \includegraphics[width=\textwidth]{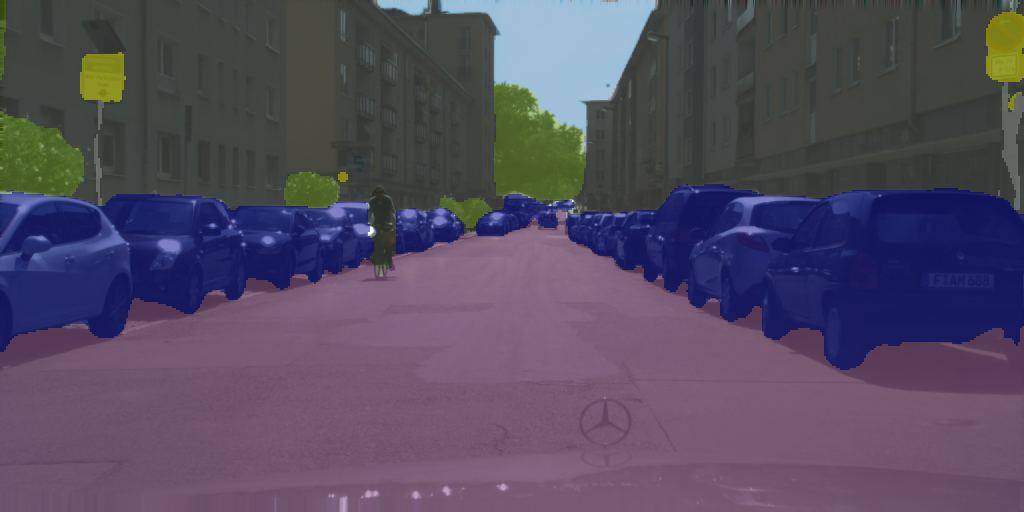}
\end{subfigure}

\begin{subfigure}{.33\textwidth}
    \includegraphics[width=\textwidth]{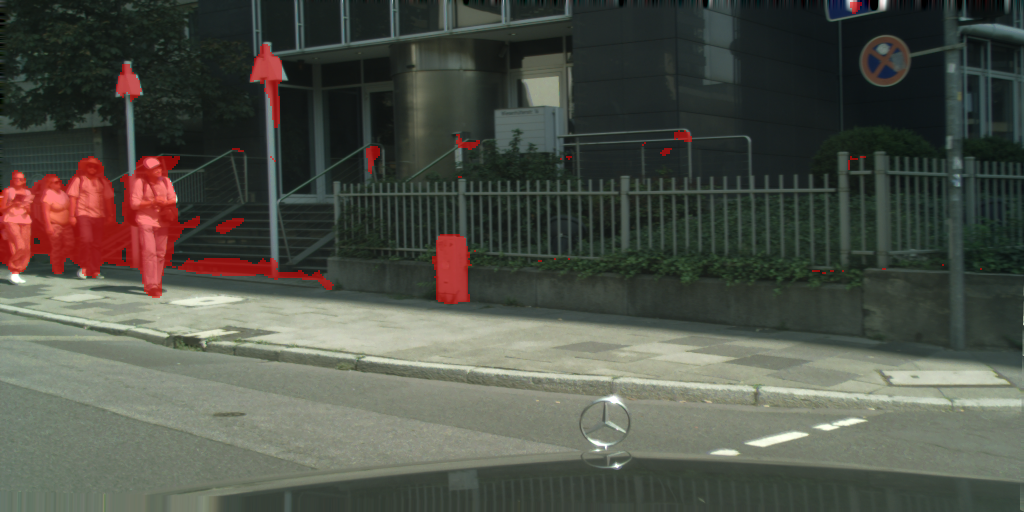}
\end{subfigure}%
\begin{subfigure}{.33\textwidth}
    \includegraphics[width=\textwidth]{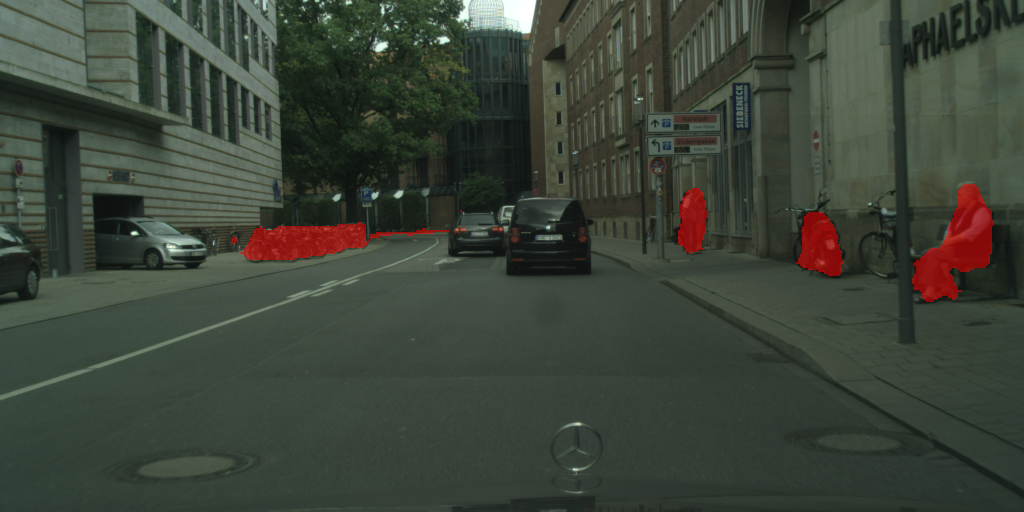}
\end{subfigure}%
\begin{subfigure}{.33\textwidth}
    \includegraphics[width=\textwidth]{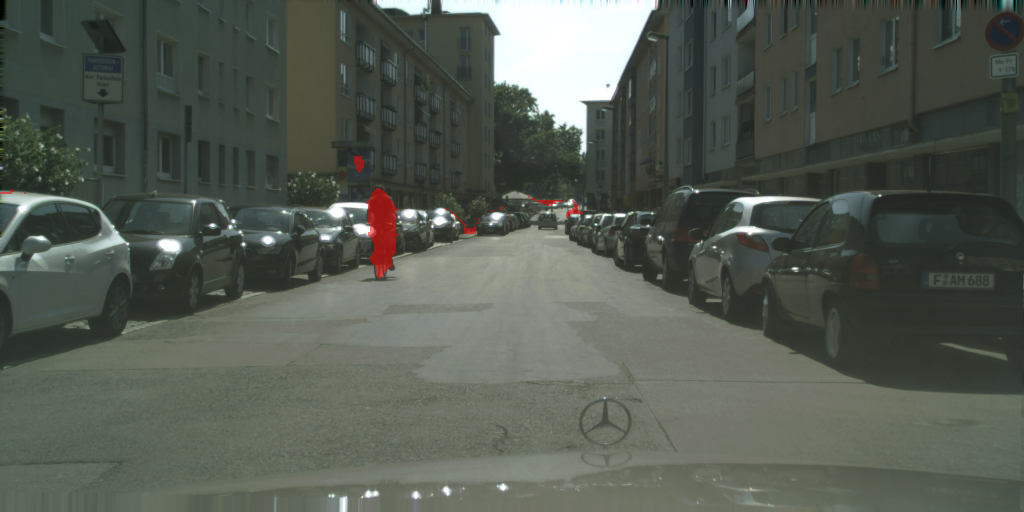}
\end{subfigure}
\caption{Predicted semantic and class agnostic segmentation on Cityscapes-VPS. Top: semantic segmentation. Bottom: class agnostic segmentation (Note: pedestrian, rider, bicycle and motorcycle are withheld from training).}
\label{fig:opensetpred_cscapesvps}
\end{figure*}

Finally, we pick the best CL variant which is performed on the prototype-level and compare to the baseline on our collected Carla dataset. Initial results, show unlike Cityscapes-VPS the baseline is not improved with the contrastive loss. There are multiple differences between both datasets regarding the size where Carla is $25 \times$ larger than Cityscapes-VPS and has more pixels labelled as unknown. Further experiments in appendix~\ref{carla_reduced} is provided to show that a reduced set of Carla will lead to the same conclusions as Cityscapes-VPS experiments. Figure~\ref{fig:opensetpred_cscapesvps} shows the results for segmenting both known classes and unknown objects on Cityscapes-VPS. It demonstrates the model ability to segment bicycle and motorcycle that was not previously seen during training. Figure~\ref{fig:opensetpred_carla} shows the results on the Carla scenarios which confirms on the model's ability to segment some of the unknown objects that did not appear during training such as in the Parking Scenario. Some of these objects Garbage Bin and Traffic Warning have distant relation to unknown objects used during the training phase as shown in Figure~\ref{fig:dendrogram}.

\begin{figure*}[!t]
\centering
\begin{subfigure}{.33\textwidth}
    \includegraphics[width=\textwidth]{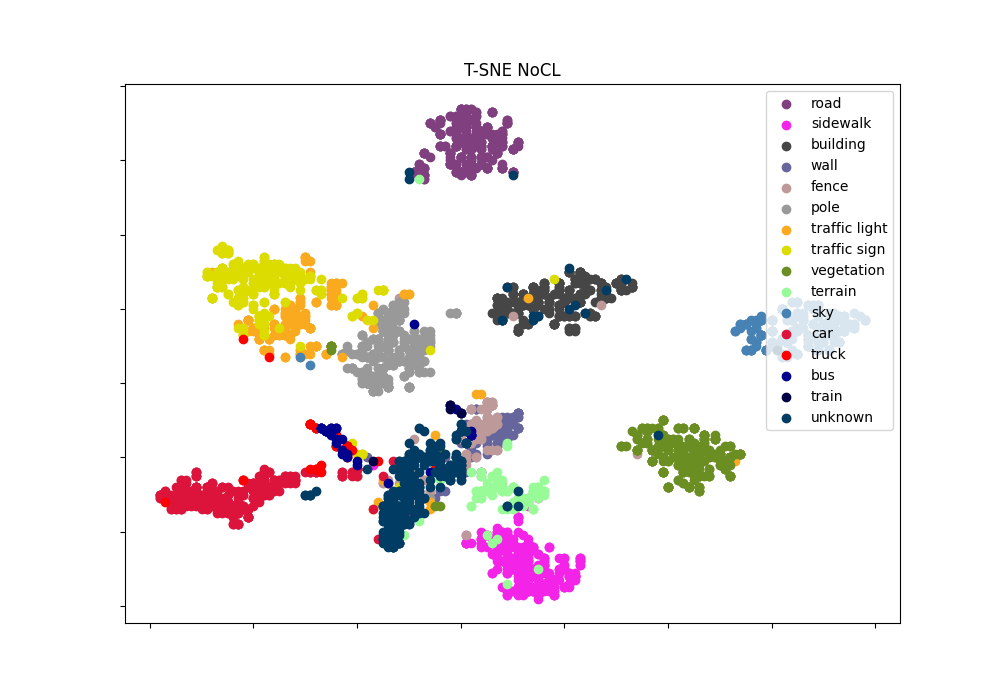}
    \caption{}
\end{subfigure}%
\begin{subfigure}{.33\textwidth}
    \includegraphics[width=\textwidth]{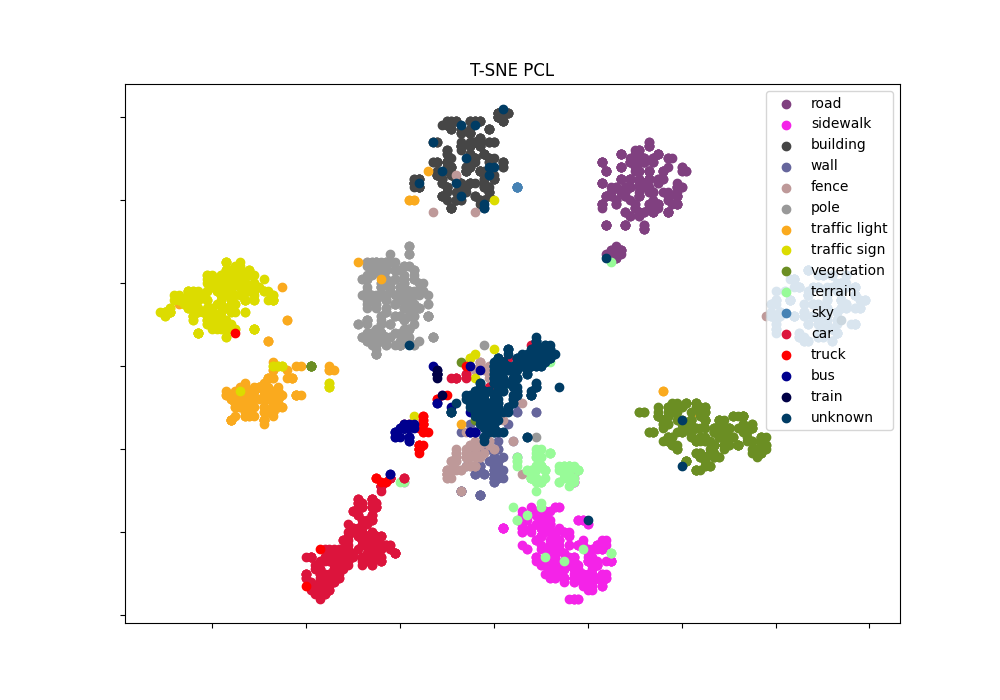}
    \caption{}
\end{subfigure}%
\begin{subfigure}{.33\textwidth}
    \includegraphics[width=\textwidth]{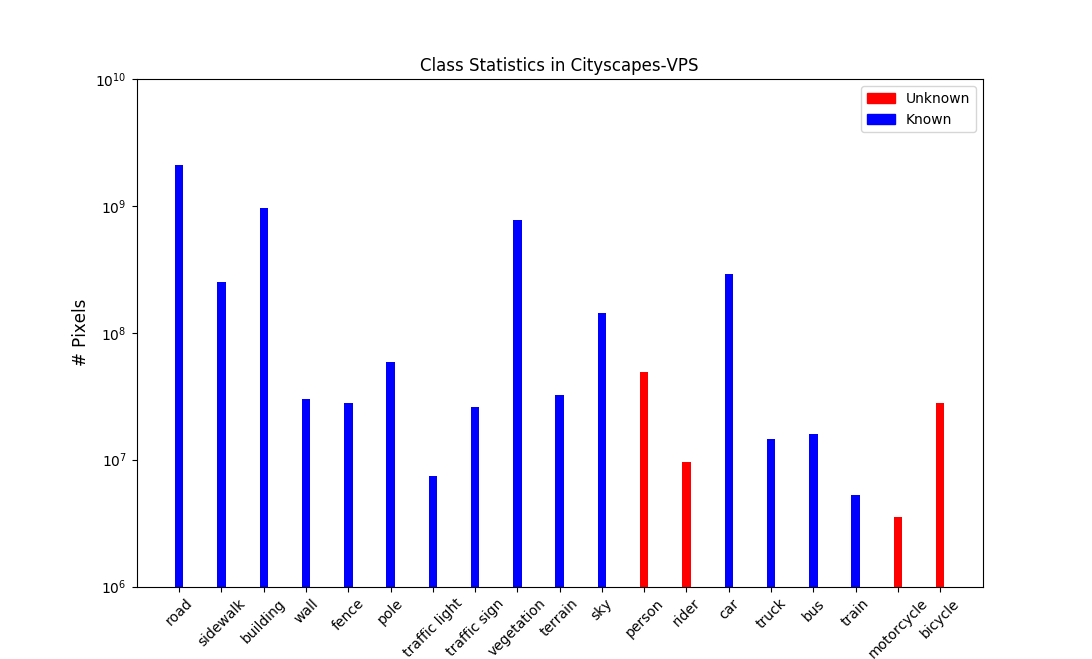}
    \caption{}
\end{subfigure}
\caption{T-SNE visualisation of the masked embeddings for 15 known classes along with the unknown objects. (a) No Contrastive Learning. (b) Prototype-level Contrastive Learning. (c) Class statistics in Cityscapes-VPS.}
\label{fig:tsne}
\end{figure*}

\subsubsection{Analysis on the Contrastive Training}
\textbf{What is the effect of the contrastive training on the shared embeddings:} Figure~\ref{fig:tsne} shows the T-SNE~\cite{van2008visualizing} visualisations for both the baseline without contrastive learning versus the contrastive learning with semantic guidance. The embeddings from the feature pyramid network are masked with the groundtruth masks of the different semantic classes to extract prototypes and gone through dimensionality reduction for visualisation. It shows how the embeddings from the contrastive learning on the prototype-level (semantic guidance) are better clustered and separated especially in severe class imbalance cases. The Figure shows the statistics per class to demonstrate which classes suffer from that. For example the baseline will lead to confusion among ``traffic sign'' and ``traffic light'', unlike the contrastive learning with better separation. It also leads to better separation of the unknown objects from the known classes such as class ``Terrian''.

\textbf{Do we need the auxiliary loss during training?} In Table~\ref{table:vca} we show the results for the baseline (No CL) but rather using pretrained weights from SimCLR~\cite{chen2020simple} versus the different contrastive learning variants. It confirms on the need to have the auxiliary loss during training of the segmentation head to improve the discrimination between known and unknown objects. Table~\ref{table:lamda} ablates the factor $\lambda$ by which we balance the main segmentation loss and the auxiliary contrastive loss. It shows generally smaller factor is better, as a factor of 1.0 degrades the segmentation of known classes. Throughout the rest of ablation experiments 0.2 is used.

\begin{table}[t]
\centering
\caption{Auxiliary loss factor for the Prototype and Temporal CL.}
\label{table:lamda}
\begin{tabular}{|l|l|c|c|}
\hline
 & $\lambda$  & mIoU & CA-IoU  \\ \hline
\multirow{3}{*}{Prototype CL} & 0.2 & \textbf{62.8} & 20.9 \\
 & 0.5 & 62.7 & 21.5 \\
 & 1.0 & 57.5 & \textbf{22.5}\\\hline
\multirow{3}{*}{Temporal CL} & 0.2 &  \textbf{61.6} &  \textbf{19.9} \\
 & 0.5 & 56.9 & 17.6\\\hline
\end{tabular}
\end{table}

\begin{table}[t]
\centering
\caption{Comparison between different Average Pooling Factors in Temporal CL.}
\label{table:gap}
\begin{tabular}{|l|c|c|}
\hline
 AP Factor  & mIoU & CA-IoU  \\ \hline
 0.05 & 61.6 & 19.9\\
 0.1 & \textbf{61.7} & \textbf{21.4} \\
 0.3 & 62.2 & 20.5\\\hline
\end{tabular}
\end{table}

\begin{table}[t]
\centering
\caption{Effect of Memory Queue in Prototype CL.}
\label{table:memory}
\begin{tabular}{|l|c|c|}
\hline
     & mIoU & CA-IoU  \\ \hline
 No Memory &  59.3 & \textbf{21.3} \\
 Memory Queue & \textbf{62.8} & 20.9 \\\hline
\end{tabular}
\end{table}

\textbf{The effect of a memory queue in contrastive learning with semantic guidance:} Table~\ref{table:memory} clearly shows the need for including a memory queue during contrastive training to increase the effective batch size for the contrastive loss. Since our encoder is a two stream model that combines depth and appearance it will not be possible to use a momentum encoder due to practical limitations in the GPU memory and compoutational resources required. Thus, using a memory queue with limited capacity is best to use in our case to ensure that only features from the latest iterations are preserved.


\textbf{The effect of average pooling in the temporal contrastive learning:} Table~\ref{table:gap} demonstrates the segmentation accuracy for both known and unknown objects with different average pooling factors. This factor is multiplied by the original feature map size to indicate the kernel size for the pooling. The smaller the factor the higher the resolution of the output. The results show that smaller factors are better as it will lead to contrasting smaller regions, this can correspond to different object parts. Unlike a larger factor of 0.3 that leads to the loss in resolution and confusion among features that can correspond to different semantic objects.




\section{Conclusion}
In this paper we study the video class agnostic segmentation task and propose novel contrastive learning variants. Our proposed contrastive learning with semantic and temporal guidance are suitable to the dense segmentation task in video sequences unlike previous contrastive learning variants. An ablation study on Cityscapes-VPS demonstrate the gain from the auxiliary contrastive loss for the main task of unknown objects segmentation. Analysis on the embeddings confirms these findings, and more insights on what is necessary in order for the auxiliary loss to improve is provided. Additionally, we provide synthetic dataset for a controlled set of experiments on large-scale video sequences. The results on our collected synthetic dataset leaves an open question on how contrastive training can benefit the video class agnostic segmentation even with large scale training data.

\appendix
\section{Appendix}

In the supplementary materials we mainly investigate three main questions: (1) What is the effect of reduced variability in the objects labelled as unknown during training? (2) What are the reasons behind contrastive learning improvement? (3) What is causing discrepancy in the results between Cityscapes-VPS and our synthetic dataset.

\subsection{Reduced Variability in Objects labelled as Unknown during Training}
\label{improved_base}
In the main experiments we initially use polynomial learning rate scheduling, since it is commonly used in state of the art methods~\cite{chen2017rethinking}. However, we further validated that using step learning rate scheduling improves the baseline with a significant margin. We show in Table~\ref{table:batch4_steplr} the results for experiments using step learning rate scheduling and larger batch size (4). It shows that contrastive learning with semantic guidance improves the CA-IoU but with a minimal gain of 0.5\%. However, more experiments on reducing the number of objects labelled as unknown demonstrate the benefit from using contrastive learning with semantic guidance on CA-IoU. In these sets of experiments only pixels belonging to class ``Person'' or originally ignored in Cityscapes are labelled as unknown during training.

\subsection{Why Prototype and Temporal Contrastive Learning Improves?}
The main reasons behind prototype and temporal contrastive learning improvement are two fold. In the prototype contrastive learning as detailed in~\cite{khosla2020supervised} the semantic guidance will lead to increased positives and negatives. This consequently leads to improvement in the discimination between signal (i.e. the prototype of a certain class) and noise (i.e. negative prototypes of others including unknown objects). In~\cite{khosla2020supervised} it was shown that the contrastive loss, whether supervised or unsupervised, learns to perform hard negative mining implicitly. It does so by increasing the gradient contribution of hard examples, and decreasing it for the easy examples. It was also shown that specifically for supervised contrastive learning, increasing the positives and negatives leads to an increase in the gradients contribution when dealing with hard positives. These two main reasons explain the improvement from prototype-level contrastive learning over the baseline, especially in case of lower variability in the unknown objects used to train the global constant. As for the temporal contrastive learning, the improvement stems from constraining the features to be temporally consistent. Since the contrastive loss samples one region as an anchor and its aligned region (i.e. warped representation using optical flow) as positive, while the remaining regions are sampled as negatives.

\begin{table}[t]
\centering
\caption{Quantitative Results on Cityscapes-VPS with larger batch size (4) and Step Learning Rate Scheduling. FD: Full Data. LU: Less number of objects labelled as unknown during training.}
\label{table:batch4_steplr}
\begin{tabular}{|l|l|c|c|c|}
\hline
 Data Mode & Method & Batch & mIoU & CA-IoU \\ \hline
\multirow{3}{*}{FD} & No CL & 4 & \textbf{63.1} & 21.0\\
&  PCL & 2 & 62.7 & \textbf{21.5}\\
&  PCL & 4 & 63.0 & 21.3\\\hline
\multirow{3}{*}{LU} & No CL & 4 & 63.1 & 18.8\\
&  PCL & 2 & \textbf{63.4} & \textbf{20.0}\\\hline
\end{tabular}
\end{table}

\begin{table}[t]
\centering
\caption{Quantitative Results on Carla with less data and less number of objects labelled as unknown. LD: Less Data (2400 frames similar to Cityscapes-VPS). LU: number of pixels labelled as unknown during training.}
\label{table:carla_2}
\begin{tabular}{|l|l|c|c|}
\hline
 Data Mode & Method & mIoU & CA-IoU \\ \hline
\multirow{3}{*}{LD + LU} & No CL  & 38.5 & 6.5\\
&  PCL & \textbf{41.7} & \textbf{16.0}\\\hline
\end{tabular}
\end{table}

\subsection{Investigating the Discrepancy between Cityscapes-VPS and Synthetic data Results}
\label{carla_reduced}
In the synthetic dataset (Carla) we found that the baseline without contrastive learning outperforms the contrastive learning with semantic guidance. So we use a reduced version of the synthetic dataset with 2400 images similar to Cityscapes-VPS dataset size. Then we reduce the pixels labelled as unknown during training to only two objects. Table~\ref{table:carla_2} demonstrates that, less data and less number of objects labelled as unknown during training, leads to clear gain from the contrastive learning with semantic guidance. This conforms with the above experiments on Cityscapes-VPS as well. In summary, our proposed auxiliary contrastive loss is more suitable and will lead to improvements when facing problems with relatively medium-scale data for the known classes and less variability in the objects labelled as unknown during training. It leaves an open question on how the auxiliary contrastive loss can improve even with abundant data. For our future work, we want to explore how to segment unknown objects without learning the global constant that was also used in~\cite{wong2020identifying}.


{\small
\bibliographystyle{ieee_fullname}
\bibliography{egbib}
}

\end{document}